\documentclass[graybox]{svmult}
\usepackage{natbib}
\usepackage[ruled,vlined]{algorithm2e}
\usepackage{multirow}
\usepackage{colortbl}
\usepackage{amsmath}
\usepackage{xcolor}
\usepackage{booktabs}

\colorlet{tableheadcolor}{gray!25} 
\colorlet{tablerowcolor}{gray!10} 
\newcommand{\rowcol}{\rowcolor{tablerowcolor}} %

\definecolor{Gray}{gray}{0.9}

\usepackage{type1cm}        
\usepackage{makeidx}         
\usepackage{graphicx}        
\usepackage{multicol}        
\usepackage[bottom]{footmisc}

\usepackage{newtxtext}       %
\usepackage{newtxmath}       

\usepackage{tikz}
\usetikzlibrary{matrix,chains,positioning,decorations.pathreplacing,arrows}

\makeindex             

\begin{document}

\title*{Artificial Neural Networks to Impute Rounded Zeros in Compositional Data}
\author{Matthias Templ}
\institute{Matthias Templ \at Institute of Data Analysis and Process Design, Zurich University of Applied Sciences, Rosenstrasse 3, CH-8401 Winterthur, Switzerland \email{matthias.templ@zhaw.ch}
}
%
%
\maketitle

\abstract*{
Generally, methods for compositional data analysis can only be applied to observed positive entries in a data matrix. 
Therefore one tries to impute missing values or measurements that were below a detection limit (rounded zeros). 
In this contribution, a new method for imputing rounded zeros based on artificial deep neural networks is shown and compared with conventional methods. 
We are also interested in the question whether the use of artificial (deep) neural networks (ANN), a representation of the data in log-ratios for imputation purposes, is relevant. 
It can be shown, that ANNs are competitive or even performing better when imputing rounded zeros of data sets with moderate size. They deliver better results when data sets are big. Also, we can see that log-ratio transformations within the artificial neural network imputation procedure nevertheless help to improve the results. This proves that the theory of compositional data analysis and the fulfillment of all properties of compositional data analysis is still very important in the age of deep learning.  
}

\abstract{
Methods of deep learning have become increasingly popular in recent years, but they have not arrived in compositional data analysis. 
Imputation methods for compositional data are typically applied on additive, centered or isometric log-ratio representations of the data. 
Generally, methods for compositional data analysis can only be applied to observed positive entries in a data matrix. 
Therefore one tries to impute missing values or measurements that were below a detection limit. 
In this paper, a new method for imputing rounded zeros based on artificial neural networks is shown and compared with conventional methods. 
We are also interested in the question whether for ANNs, a representation of the data in log-ratios for imputation purposes, is relevant. 
It can be shown, that ANNs are competitive or even performing better when imputing rounded zeros of data sets with moderate size. They deliver better results when data sets are big. Also, we can see that log-ratio transformations within the artificial neural network imputation procedure nevertheless help to improve the results. This proves that the theory of compositional data analysis and the fulfillment of all properties of compositional data analysis is still very important in the age of deep learning.  
}

\section{Introduction}
\label{sec:1}

\noindent \textit{Compositional data and rounded zeros:} 
Compositional data are quantitative descriptions of the parts of some whole, whereby the important information is not the absolute values but relative information. Measurements involving proportions, percentages, probabilities, concentrations are compositional data.
An example of compositional data is the 24 hours day categorized into sleeping time, working time, lunch break, and free time. There exists inner dependencies: if one sleeps longer at a day, automatically less time for other things is left.
Also not all values are possible (e.g. a day is restricted to 24h) and the compositions are represented in the simplex instead of an Euclidean geometry. As for the previous example, the sleeping time plus working time cannot be larger than 24 hours, for example. 
Statistical methods are mostly designed to be applied in the Euclidean geometry. A log-ratio presentation of the data is a way out to shift the original measurements from the simplex to the Euclidean geometry. 
However, zeros or rounded zeros are problematic when computing log-ratios. Either in the denominator, the ratio is not defined or in the numerator the log is not defined. 
Zeros are true values where the measurement reports zero. A rounded zero occurs whenever a measurement unit cannot measure a too small amount in a sample. For example, a measurement unit can only detect ppm's of a certain chemical element higher than a certain threshold - the detection limit - and reports zeros whenever there are fewer ppm's in the sample. The value is thus rounded to zero, but only because the measurement unit could not detect anything even though we know that there must be a certain contribution of this chemical element in the sample. Values below the detection limit are thus known under the name rounded zeros \citep{Aitchison86,martin03} and represent left censored data. 

It is fully meaningful to impute rounded zeros with a reasonable small value and continue with the processing of a complete data set \citep{Filzmoser_2018}. To guarantee imputation by a considerable small value, an upper bound should be considered - typically the detection limit of this variable. Any imputation method should then impute with values below this upper bound, but larger than zero. 

To give an example as outlined also similarly in \cite{Filzmoser_2018}, a rounded zero value present for a variable, say \textit{Arsenic} and the detection
limit for \textit{Aresenic} is 0.1 mg/kg. An imputed value should 
be thus in the interval $(0,0.1)$. The imputation should consider the information about all other compositional parts, i.e. a multivariate approach is preferable. 
Furthermore, the log-ratios between other 
parts should remain unchanged after the imputation of the rounded zeros.




The imputation of rounded zeros can be considered as a very difficult problem with different solutions proposed in the literature \citep{martin03, martin07, Palarea08, Martin12, palarea13, palarea14, martin15cz, boogaart15, templ16highdim, chen17}. More on this in Section~\ref{rounded}. \\

\noindent \textit{Artificial deep neural networks:} 
A neural network is just a non-linear statistical model \citep{hastie09}, which is based on weighted linear combinations of the sample values.
Within an ANN, the aim is to find millions of weights to get the best possible output from input data and multiple layers. 
The weights are updated iteratively. In the first iteration, the weights are random and the result is accordingly bad. For each iteration, we go in the  ``optimum direction'', whereby a gradient method is used for this. It requires backpropagation with an optimizer (e.g. Adam) and a loss function (e.g. mean squared error). The quality of the predictions is evaluated based on a selected metric (e.g. mean absolute error) on validation and training data. 

Often missing values are just initialized before the neural network is fitted \citep{smieja18}.
Also, missing values may be incorperated in a classification problem without imputing them. The trick is to add an indicator matrix determining the position of missing values to the data matrix and then using a generative additive network (GAN). For more details, see \cite{misgan19, gan18, mattei18}. Note that this is applied within several image inpainting methods \citep[see, e.g.,][]{xie12}.
However, our aim is different. Rounded zeros are imputed with fixed values so that the imputed data set can be used by any researcher or practitioner for any analysis that needs complete data. \\

\noindent \textit{ANNs for missing values:} 
Artificial neural networks have been rarely used for the imputation of missing values purposes. \cite{maiti08} used neural networks for an agriculture survey (and compared it with hot-deck imputation). Various other contributions took place around 2007/2008.
Afterwards, \cite{jerez10} used a multi-layer perceptron method for the imputation of breast cancer data. \cite{mccoy18} used variational autoencoders for imputation.
\cite{choud19} and \cite{silva15} used artificial neural networks. Another approach is to not explicitly impute missing values, but to add the missingness matrix. Hereby, a position of the original data matrix with missing value is denoted with a 1, and non-missing values are denoted with 0. This matrix is then integrated  into the whole model. \cite{misgan19}, \citep{gan18} and \citep{mattei18} used generative artificial networks for classification purposes under missing values. For special applications, these were also used by \cite{lim19} and \cite{aris19} (RNA data). However, no ANN was used in the context of compositional data analysis and in the context of the imputation of rounded zeros in compositional data. \\

\noindent \textit{Outline:} 
In Section, \ref{rounded} methods for the imputation of rounded zeros are listed and Table~\ref{tabmethods} summarizes them. 
Section~\ref{ann} introduces the specific type of artificial neural networks that is relevant in the context of imputation for compositional data.
The parameters of our imputation procedure that have been made available in software are described in Section~\ref{annimp} for the imputation of rounded zeros in one variable. The focus of Section~\ref{em} is the adaptation to impute rounded zeros as an Expectation-Maximum (EM) method. This allows also to impute not only one variable at a time but sequentially all compositional parts with rounded zeros in a data set. 
Two different approaches are presented to replace rounded zeros in compositional data, one using a log-ratio transformation and the other without using a log-ratio representation.  
The outcome and comparison of the methods on real-world data sets are shown and described in Section~\ref{numeric}. Section~\ref{conclusions} concludes and summarizes the major findings. 

\section{General imputation methods and imputation methods for rounded zeros}
\label{rounded}

Note that the simplest method is univariate and imputes zeros with 65\% of the detection limit ($DL$). This minimizes the distortion of the covariance structure \citep{martin03,MPO2011}. However, this kind of mean imputation leads to an underestimation of the compositional variability. 
The variability can be increased by drawing random numbers from a uniform distribution in $(0,DL)$, with DL, the detection limit of a variable, but then the precision of the imputations lowers. 
Alternatively, \cite{palarea13} and \cite{palarea14} used a lognormal distribution, truncated by the 
threshold. In any case, all these methods represent univariate approaches to a multivariate compositional problem. Note that the methods from \cite{palarea13} and \cite{palarea14} needs at least one fully observed variable. 
In our application, we compare only multiple and multivariate methods and leave out univariate methods. 

EM-based model-based methods offer multivariate approaches for the imputation of rounded zeros and by using censored regression it can be guaranteed that imputed values do not exceed a defined threshold. A modified EM algorithm \citep{martin07,Palarea08} in additive log-ratio coordinates \citep{Aitchison86} was reformulated by using pivot log-ratio coordinates \citep{Filzmoser_2018} in \cite{Martin12}. Both algorithms uses censored multiple regression for each variable in an EM-based manner until the imputations stabilize. 
The algorithm is complex because detection limits must also be represented in isometric log-ratio coordinates and inverse isometric log-ratio transformation must happen in each step of the algorithm together with an adjustment of values to not lose the absolute information of the original data. This is necessary when we still want to know, for example, the exact household expenditures in expenditure data and not only the ratios between them. A detailed description of this complex algorithm can be found in \cite{Martin12, Filzmoser_2018}. The method itself needs also an initialization of rounded zeros (for example using the 65\% method) because otherwise log-ratios cannot be calculated from zero values. Instead of ordinary least squares regression, a robust estimator can be plugged in. 

For high-dimensional data, two other methods to replace rounded zeros should be mentioned. 
\cite{templ16highdim} uses partial least squares regression for imputing all variables sequentially and repeatedly in an EM-based algorithm. To decrease computation time, \cite{chen17} uses a clustering approach to select variables within the algorithm. 

In addition to methods suitable to handle rounded zeros by imputing below a given detection limit, we also want to compare popular imputation methods that do not take these limits into account. 
A major point of criticism of this contribution could be that methods for rounded zeros are compared with methods that do not take into account the special problem of rounded zeros. The motivation for this comparison comes from practice and is not scientifically based, but is based on the experience as package maintainer of VIM \citep{templ12VIM, VIMJSS}, and robCompositions \citep{templ11vera, Filzmoser_2018} - both containing imputation methods. This experience says that readers need to see a comparison to understand that only (compositional) methods that take the problem of rounded zeros into account successfully impute rounded zeros. Up to our knowledge, articles on rounded zero imputation have only compared specific methods for rounded zeros, with the result that users without the obvious comparison still resort to unsuitable methods, because they are just trendy to use for more general imputation problems. 
Many users are also confused about whether they should use imputation methods for missing values in compositional data or methods for replacing rounded zeros. We show what happens when the wrong methods are selected and what the consequences are. 
Therefore, methods for rounded zeros of compositional data are also compared to very popular imputation methods: imputation using random forests implemented in R package \texttt{missForest} \citep{Stekhoven11} or the faster implementation in \texttt{missRanger} \citep{missRanger}, predictive mean matching using R package \texttt{mice} \citep{mice}, $k$-nearest neighbor imputation of R package \texttt{VIM} \citep{VIMJSS}, missing value imputation using pivot coordinates as implemented in R package \texttt{robCompositions} in function \texttt{impCoda} \citep{hron10} and from the same package a $k$-nearest neighbor method for missing values using Aitchison distances in function \texttt{kNNa} \citep{hron10}.

\begin{table}[!htp]
\begin{small}
\caption{\label{tabmethods}Methods compared in this contribution. Univariate methods were not considered. We distinguish between methods for the imputation of missing values (not designed for detection limit problems) and rounded zeros (considering detection limites) and methods that do or do not take the compositional nature of compositional data into account.}
\begin{tabular}{lp{3.5cm}|p{6cm}}
\toprule
Method & Reference & Description \\
\midrule
\rowcol
\multicolumn{3}{l}{Non-compositional approaches not designed for detection limit problems (non-CoDa, non-DL)} \\
deepImp & \cite{deepImp} & ANNs in an EM-based implementation (serves as a benchmark for deepImp-dl) \\
kNN & \cite{VIMJSS} & $k$ nearest neighbor imputation \\
mice & \cite{mice} & Predictive mean matching in an EM-based implementation \\
missForest (rf) & \cite{Stekhoven11} & Imputation using random forests in an EM-based implementation \\
\midrule
\rowcol
\multicolumn{3}{l}{Non-compositional approaches considering detection limits (non-CoDa, DL)} \\
deepImp-dl & \cite{deepImp} & ANNs in an EM-based implementation considering detection limits (serves as a benchmark for deepImpCoDa-dl)\\
 \midrule 
\rowcol
\multicolumn{3}{l}{Compositional methods ignoring not designed for detection limit problems (CoDa, non-DL)} \\
deepImpCoDa & this article & ANNs in an EM-based implementation considering compositinal data (serves as a benchmark for deepImpCoDa-dl) \\
aknn & \cite{hron10}  & $k$NN imputatation for compositional data \\
impCoDa & \cite{hron10}  & EM-based regression imputation using pivot coordinates. \\
\midrule
\rowcol
\multicolumn{3}{l}{Compositional methods considering detection limits (CoDa, DL)} \\
deepImpCoDa-dl & this article & ANNs in an EM-based implementation considering detection limits \\
imputeBDL-lm & \cite{templ16highdim}  & EM-based regression imputation using pivot coordinates and considering detection limits. \\
imputeBDL-pls & \cite{templ16highdim}  & EM-based partial least-squares regression imputation using pivot coordinates and considering detection limits. \\
imputeBDL-rob & \cite{templ16highdim}  & EM-based robust (MM) regression imputation using pivot coordinates and considering detection limits. \\
impRZilr & \cite{Martin12}  & EM-based partial least squares regression imputation using pivot coordinates and considering detection limits. \\
\bottomrule
\end{tabular}
\end{small}
\end{table}


\section{Artificial neural networks}
\label{ann}


In this section the general construction of a neural network is explained, details about the neural network used for the imputation are given in the next section. 

Neurons hold some information on each variable and a set of observations. Each neuron has an activation, depending on how the input information looks like. A layer in a deep neural network is a collection of neurons in one step. There are three types of layers: the input layer, the hidden layers, and the output layer. A neuron $\nu_1^l$ in the input layer, for example, takes the observations of a data set as activations and the hidden layers represent non-linear weighted observations. 

The activations $\mathbf{a}$ should take values between zero and one. 
The (default) activation function reLU (rectified Linear Unit) \citep{relu} takes the value zero if the activation is negative. Most activation functions use a non-linear function mapping between the inputs and the response. 


In the first hidden layer $\iota_{l}$ with $l = 2,...,\Phi-1$ with each neuron $\nu_{v}$ with $v = 1,..,\phi_l$ represents very detailed information, while passed through the layers, the information gets generalized and in the end summed up to either one or multiple neurons depending on the type of variable. $\Phi$ represents the number of layers, including the input and the output layer, and $\phi$ the number of neurons within the layer $l$. Figure~\ref{layers} shows a simplified illustration of a network with two hidden layers and few neurons. 

\def\layersep{2.5cm}

\begin{figure}[!htp]
\begin{tikzpicture}[
   shorten >=1pt,->,
   draw=black!50,
    node distance=\layersep,
    every pin edge/.style={<-,shorten <=1pt},
    neuron/.style={circle,fill=black!25,minimum size=17pt,inner sep=0pt},
    input neuron/.style={neuron, fill=green!50},
    output neuron/.style={neuron, fill=red!50},
    hidden neuron/.style={neuron, fill=blue!50},
    annot/.style={text width=4em, text centered}
]

    \foreach \name  [count=\y] in 
    {$\mathbf{x}_1$, $\mathbf{x}_2$, $\mathbf{x}_3$, $\mathbf{x}_4$, $\mathbf{x}_5$, $\mathbf{x}_6$, $\mathbf{x}_7$, $\mathbf{x}_8$, $\mathbf{x}_n$}
     {   \node[input neuron, pin=left:\name] (I-\y) at (0,-\y cm) {};  
     }

    \newcommand\Nhidden{2}
    \newcommand\NodOne{11}
    \newcommand\NodTwo{8}
    \newcommand\Nod{9}


     \foreach \y in {1,...,\NodOne} {
          \path[yshift=1.80cm]
              node[hidden neuron] (H1-\y) at (1*\layersep,-\y cm) {};
              }
    \node[annot,above of=H1-1, node distance=1cm] (hl1) {Hidden layer $\iota_{2}$};
     \foreach \y in {1,...,\NodTwo} {
          \path[yshift=0cm]
              node[hidden neuron] (H2-\y) at (2*\layersep,-\y cm) {};            
           }
    \node[annot,above of=H2-1, node distance=1cm] (hl2) {Hidden layer $\iota_{3}$};          


    \node[below=0.5cm of H1-\NodOne] (text1) {no. of neurons: $\phi_2$};
    \node at (text1 -| H2-1) (text2) {no. of neurons: $\phi_3$};

    \node[output neuron,pin={[pin edge={->}]right:Output}, right of=H\Nhidden-5] (O) {};

    \node at (text1 -| O) (text3) {};
    \foreach \source in {1,...,9}{
        \foreach \dest in {1,...,\NodOne}{
            \path (I-\source) edge (H1-\dest);
         }
    }
    \foreach [remember=\N as \lastN (initially 1)] \N in {2,...,\Nhidden}
       \foreach \source in {1,...,\NodOne}
           \foreach \dest in {1,...,\NodTwo}
               \path (H\lastN-\source) edge (H\N-\dest);

    \foreach \source in {1,...,\NodTwo}
        \path (H\Nhidden-\source) edge (O);


    \node[annot,left of=hl1] {Input layer $\iota_{1}$};
    \node[annot,right of=hl\Nhidden] {Output layer $\iota_{\Phi}$};
\end{tikzpicture}
\caption{\label{layers}Simplified schematic overview of a neural network with a few observations, two hidden layer with 13 and 9 neurons. Note that the default deep artificial neural network consists of $n$ observations, 10 layers, 1000 neurons in the first hidden layer, 900 in the second, ..., 100 in the last hidden layer and the output depends on the scaling of the target variable. In total 3325601 parameters (mostly weights) are trained in the default network.}
\end{figure}
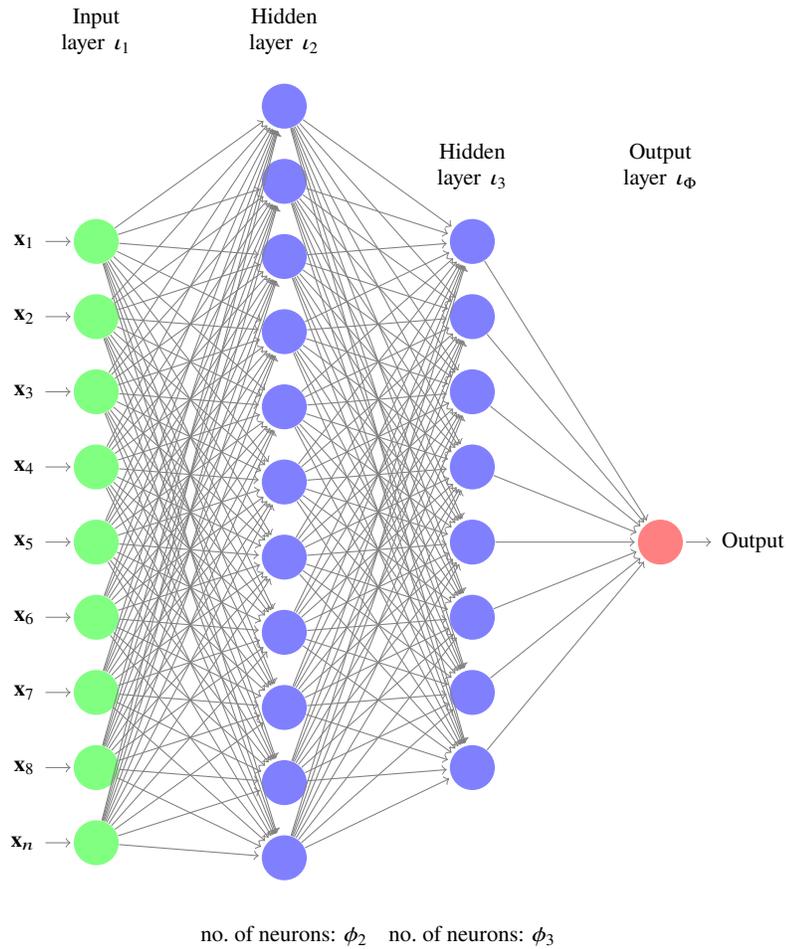

%

\begin{figure}[!htp]
\begin{tikzpicture}[
init/.style={
  draw,
  circle,
  inner sep=2pt,
  font=\Huge,
  join = by -latex
},
squa/.style={
  draw,
  inner sep=2pt,
  font=\Large,
  join = by -latex
},
start chain=2,node distance=13mm
]
\node[on chain=2] 
  (x2) {$x_2$};
\node[on chain=2,join=by o-latex] 
  {$\omega_2$};
\node[on chain=2,init] (sigma) 
  {$\displaystyle\Sigma$};
\node[on chain=2,squa,label=above:{\parbox{2cm}{\centering Activation \\ function}}]   
  {$f$};
\node[on chain=2,label=above:Neuron,join=by -latex] 
  {$v_1^2$};
\begin{scope}[start chain=1]
\node[on chain=1] at (0,1.5cm) 
  (x1) {$x_1$};
\node[on chain=1,join=by o-latex] 
  (w1) {$\omega_1$};
\end{scope}
\begin{scope}[start chain=3]
\node[on chain=3] at (0,-1.5cm) 
  (x3) {$x_3$};
\node[on chain=3,label=below:Weights,join=by o-latex] 
  (w3) {$\omega_3$};
\end{scope}
\node[label=above:\parbox{2cm}{\centering Bias \\ $\beta_1$}] at (sigma|-w1) (b) {};

\draw[-latex] (w1) -- (sigma);
\draw[-latex] (w3) -- (sigma);
\draw[o-latex] (b) -- (sigma);

\draw[decorate,decoration={brace,mirror}] (x1.north west) -- node[left=10pt] {Inputs} (x3.south west);
\end{tikzpicture}
\caption{\label{neurons}This schematic illustration tells the story behind the calculation of one single neuron with respect to the first hidden layer. This neuron serves then as input for the next layer.}
\end{figure}
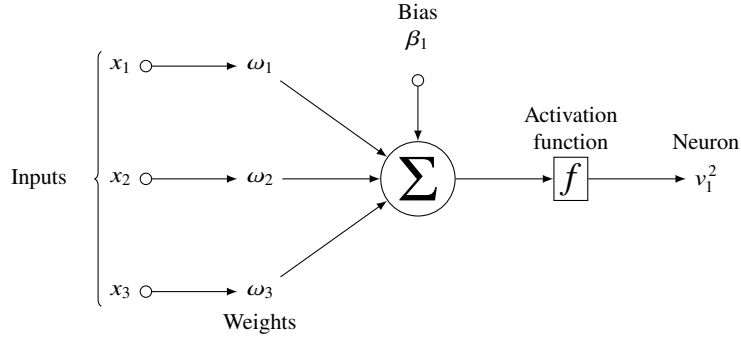

The goal of training a network is to find the optimal weights $\omega_{\nu_v^l, \nu_{v}^{l+1}}$ for each connection between the neurons of two layers (see also Figure~\ref{neurons}). Then the activations of the neurons from the first layer are taken and the weighted sum according to these weights is computed.

Let $\boldsymbol{\Omega}$ be the matrix of weights 
in each layer.
The weights of the $l$-th layer are defined as $\boldsymbol{\omega}^l$. The activation values of the neurons in one layer are $\boldsymbol{\nu}^l = \nu^l_1, ... ,\nu^l_{\phi_l}$ while the biases within one layer are $\boldsymbol{\beta}^l = \beta_1^l, ... ,\beta_{\phi_l}^l$. The formula for the transition of one layer to the other is then defined by
$relu(\boldsymbol{\Omega} \mathbf{\nu}^l + \boldsymbol{\beta})$. 

After setting initially all weights randomly, a loss function of the network is defined. The output of the loss function is a single number judging the quality of the neural network. To lower the value of the loss function, an adaptive moment estimation called Adam \citep{adam,Ruder16} is used (other methods can be selected), which is a stochastic gradient descend method that uses adaptive learning rates for each parameter of the algorithm. 
With this gradient descend optimization all the weights ${\boldsymbol{\Omega}}$ are optimized to reach the next local minimum $-\nabla C({\boldsymbol{\Omega}})$ resulting in the most rapid decrease. 
This is causing the most rapid decrease in our loss function.
The gradient used to adjust these weights is computed with back propagation, whereby the weights gets updated. 

The weighted sum, which defines the activation of a neuron is defined by $\boldsymbol{z}^l = \boldsymbol{\omega}^l\boldsymbol{\nu}^{l-1}+\boldsymbol{\beta}^l$. The activation of a neuron $\boldsymbol{\nu}^l$ in layer $l$, is then defined with $\boldsymbol{\nu}^l = relu(\boldsymbol{z}^l)$. The loss is therefore computed by comparison of $\boldsymbol{z}^l$ and $\boldsymbol{\nu}^l$. The sensitivity of the loss function $C$ with regard to changes in the weights $\boldsymbol{\omega}^l$ is calculated with the chain rule $$\frac{\partial{C}}{\partial \boldsymbol{\omega}^l} =\frac{\partial\boldsymbol{z}^l}{\partial\boldsymbol{\omega}^l} \frac{\partial\boldsymbol{\nu}^l}{\partial\boldsymbol{z}^l} \frac{\partial{C}}{\partial\boldsymbol{\nu}^l} \quad .$$
This means that the amount of this nudge influencing the last layer, depends on the strength of the previous neuron, $\frac{\partial\boldsymbol{z}^l}{\partial\boldsymbol{\omega}^l}=\boldsymbol{\nu}^{l-1}$.

This is done over all training observations, $\frac{\partial{C}}{\partial\boldsymbol{\omega}^l} = \frac{1}{n}\sum^{n-1}_{k=0}(\frac{\partial{C}_k}{\partial\boldsymbol{\omega}^l})$. 
The gradient vector $\nabla{C}$ contains all the partial derivatives from the loss function with respect to all weights and biases.
The sensitivity of the loss function to changes in the bias is computed equivalent with $$\frac{\partial{C}}{\partial \boldsymbol{\beta}^l} = \frac{\partial\boldsymbol{\nu}^l}{\partial\boldsymbol{z}^l} \frac{\partial{C}}{\partial\boldsymbol{\nu}^l} \quad .$$ 

The above described approach is then applied and computed through the complete neural network. The weights will be adjusted and the computation is iterated and repeated many times (also called the number of epochs, see below). More details about how neural networks are computed can be found, for example, in \cite{Nielsen15}.


\section{Artificial neural networks to impute rounded zeros - parameter settings}
\label{annimp}

In the following an artificial neural network is described and adapted for imputation purposes, implemented in function \textit{impNNetCoDa} in R package \textit{deepImp}. This deep learning method works with a neural network based on \textit{keras} \citep{Allaire18,chollet2015keras} and \textit{tensorflow} \citep{tensorflow,tensorflow2015whitepaper}. 

Before describing the whole EM-based algorithm that automatically imputes all rounded zeros in all variables of a data set, the parameter choices available in our implementation are outlined. \\

\noindent \textit{Initialization of rounded zeros:} In principle, any other method from before can be used to initialize the rounded zeros and several choices are available in our implementation. The default approach is to use a $k$ nearest neighbor approach that uses Aitchison distances and robust means \citep{hron10}. \\

\noindent \textit{Iterations:} Initialized missing values are updated step by step with imputations. Whenever all variables with rounded zeros are updated once, one iteration has passed. This represents an outer loop of the algorithm. If the EM-based sequential algorithm does not converge after a specified number of  \texttt{iterations}, then the algorithm is stopped and a warning is supplied. \\


\noindent \textit{Threshold for convergence:} 
Given the $r$-th iteration, a change in the imputed values compared to the previous iteration is measured. More precisely, the algorithm stops as soon as the sum of squared relative distances between the imputed values in the $r$-th step and the $r-1$-th step is below the specified threshold.  \\

\begin{figure}[!htp]
\begin{center}
\includegraphics[width = 1\textwidth]{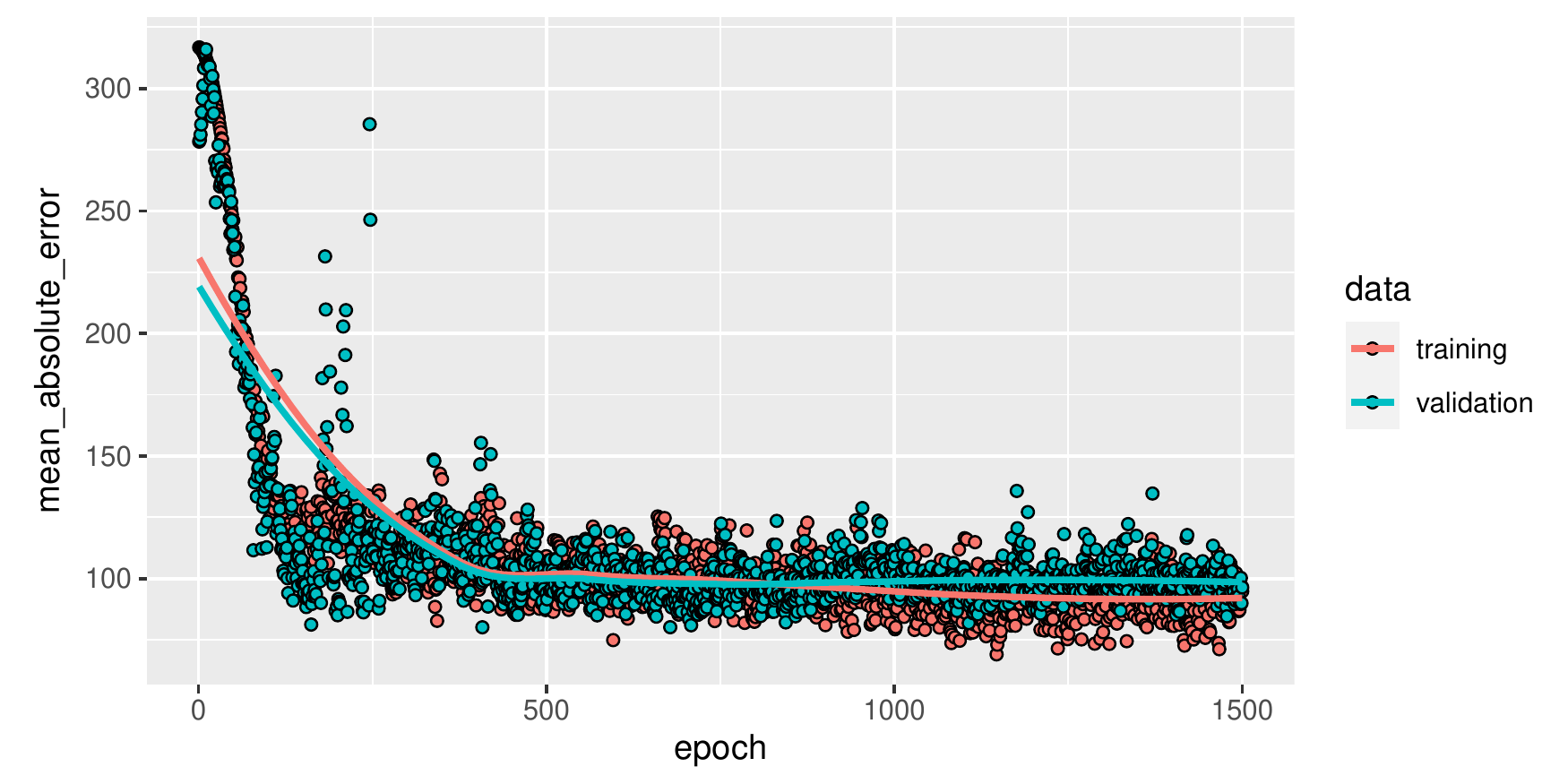}
\caption{\label{valerror1}Training and validation error for the imputation of variable Ag in the Moss data.}
\end{center}
\end{figure}

\noindent \textit{Number of epochs:}
The data set is not only sent once through the deep neural network but many times (number of epochs).
This is needed to optimize the learning, which is an iterative process. Updating the weights in one pass (i.e. one epoch) is too little. Generally, it applies that too few epochs often leads to an underfit while too many epochs can lead to an overfit. 
The optimum number of epochs is data set dependent. In all our experiments, a choice of about 200-300 epochs gave the most promising results, when either looking at and comparing validation and training errors or when evaluating it directly on the validation criteria defined in Section~\ref{val}. If the validation error is higher than the training error, this is an indication of overfitting. Figure~\ref{valerror1} corresponds to the imputation of rounded zeros in the variable Ag in the Moss data set (see Section~\ref{datasets}) using our artificial deep learning network with 10 layers and different number of epochs. No overfit is visible. 
Note that in our implementation 20\% of the observations used as the validation set by default (for each epoch). 

The number of epochs is one parameter of our algorithm with a sensible default. However, it is better to set the number of epochs to a high value instead of too small value, because in any case, the next parameter provides a stopping rule to stop the training of the deep neural network earlier.

For any given data set, the training and validation errors can be used to evaluate the correct number of epochs, but also to evaluate the number of layers and the number of neurons in the layers. If training and validation errors are quite different, too few layers, neurons, and/or epochs were selected. 
In Figure~\ref{valerror1} we see a perfect picture where both errors in the deep neural network, using the evaluation criterion \textit{mean\_absolute\_error}, are about the same size.
The optimal number of epochs is about 350, after that there is no real improvement.

Regarding imputation, it is also interesting to look at an imputation error, as long as you know the true values (as in the numerical examples in Section~\ref{numeric}). The imputation error over the validation measure CED (see Section~\ref{val}) is lowest at about 200-300 epochs for all later used data. \\

\begin{figure}[!htp]
\begin{center}
\includegraphics[width = 1\textwidth]{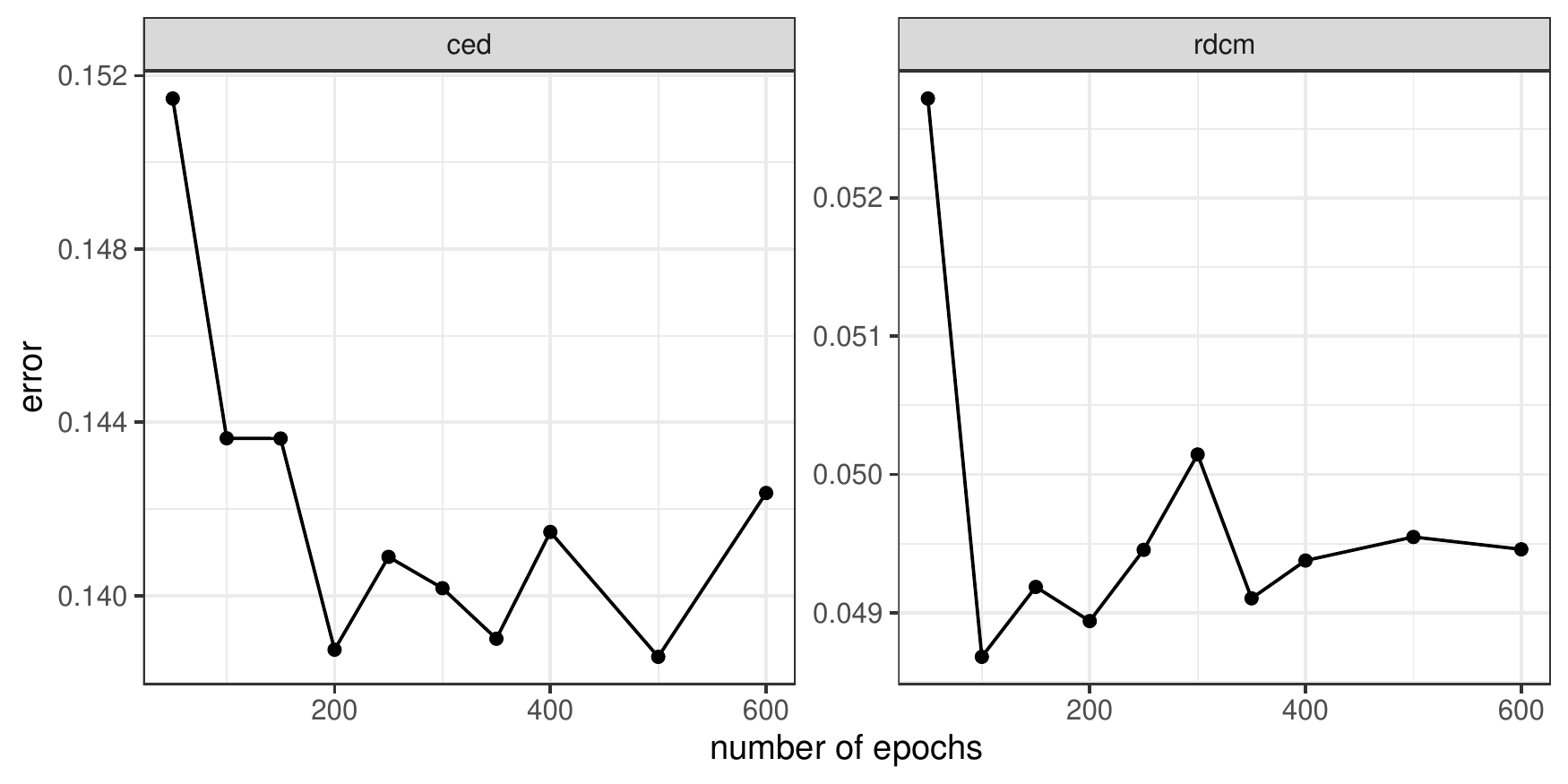}
\caption{\label{valerror2}Imputation error for different numbers of epochs for the imputation of variable Ag in the Moss data set. For our choice of the number of epochs for the numerical results section is based on the CED criteria (see Section \ref{val}).}
\end{center}
\end{figure}

\noindent \textit{Patient value:} 
The aim is to stop at those number of epochs without any significant change of weights in the last epochs (patient value). Setting this value appropriately overcomes overfitting when specifying the number of epochs too high. For example, when the weights do not change for about 25 epochs (default), the training of the weights in the deep neural network would stop. \\

\noindent \textit{Drop out:}
This is the rate of input units in each layer, which are excluded in the neural network. 
The main purpose of using drop-out is to protect against over-matching, but also to achieve different results when imputing again if the random number generator is not set to the same value. This is also the only way to go in the direction of multiple imputation. 
Although proper multiple imputation (with correct coverage rates) is not possible to achieve in general using drop-out. \\

\noindent \textit{Number and density of layers in the deep network:} 
The optimal number of layers is data-dependent. In general, too many layers can cause an overfit, while too few layers can cause an underfit. Generally, it is better to add too many layers than too few. This is the reason why by default the number of layers in the deep neural network is set to 10. 

The user can evaluate the optimal number of layers using cross-validation. The validation error is not very different from training terror, both errors are obtained from cross-validation. \\

\noindent \textit{Number of neurons in each layer:} 
The optimal number of neurons in each layer depends on your data particularities. By default, the first hidden layers consist of 1000 neurons and then decrease by 100 in each following layer so that the 10th layer has 100 neurons. Note that the user can overwrite this default choice. Again cross-validation using different choices of neurons can give some hints about the optimal choice of the number of neurons.\\

\noindent \textit{Optimizer:} 
The choice of the optimizer is crucial. It defines the gradient method used to iteratively update the weights in the neural network. Our default choice is Adam \citep{adam}, 
a first-order stochastic gradient-based optimization based on adaptive estimates of lower-order moments.

Other optimizers selectable are SGD (Stochastic gradient descent optimizer), RMSprop, Adagrad, Adadelta, Adamax, and Nadam (Nesterov Adam optimizer). Note that all methods have sensible defaults, e.g. on the learning rate. \\

\noindent \textit{Loss function:} 
The choice of the loss function depends on the scale type of the response variable (the variable to be imputed). For nominal variables to be imputed, typically a cross-entropy is used, while for continuous variables a mean squared error is used. Note that we did not replace this with a compositional counterpart. This could be future work which would need careful implementation in software. Please also note that no robust loss function can be integrated into Keras, because no quantiles or any robust measure can be used to construct a robust loss measure. Also, the deep neural networks are computationally expensive, thus replacing a mean squared error with robust measures, the computation time would increase heavily. As a consequence, all deep neural networks are non-robust and sensitive to outliers. \\

\noindent  \textit{Evaluation metrics:}
An evaluation metric is used to evaluate the performance of your model. A metric function is thus similar to a loss function. The latter is used for training the model while the evaluation metrics are used to judge the performance of the model. One can use any of the loss functions as a metric function.
In our software implementation, the default evaluation metric is the mean absolute error.\\

\noindent \textit{Activation:} 
Especially for compositional data, a good choice of an activation function is important, because the weights for neurons get transformed by the activation function. For example, when using the activation \textit{linear}, the outcome of the neural network for a 2-dimensional continuous data set in a regression context  is approximately the same as for least squares regression. 
To allow for non-linear combination of neurons, we choose by default the activation \textit{reLU}, see \cite{Vedaldi15}. 
One positive aspect of the reLU function is that it accelerates the convergence of stochastic gradient descent compared to other activation functions and it is computationally cheap \citep{Krizhevsky12}.
Other choices are, e.g., softmax, elu, selu, or sigmoid, to just name a few. For the last layer (that actually do the imputation) a different choice of activation must be taken, namely linear. 


\section{Imputation of rounded zeros using artificial neural networks in an Expectation-Maximum style} 
\label{em}

The proposed approach uses one artificial network per variable to be imputed.
Note that the structure of the output layer, the activation function, the loss, and evaluation function for the output layer depend on the distribution of a variable to be imputed. 
Naturally, this is a log-ratio, but in principle also external variables that are non-compositional could be integrated. Generally, for continuous measurements and log-ratios, the activation function relu \citep{relu} is used concerning all layers (default), the mean absolute error is used as a loss function (default) and the output layer consists of a single neuron and linear activation.

As already mentioned, the initialization of rounded zeros is necessary whenever artificial neural networks are applied to log-ratio representations instead of raw original values. 

The algorithm works in a chain, i.e. it iterates sequentially applied to all variables until convergence (EM-based method).


All networks have $\Phi = 10$ layers (by default), which in experimental pre-studies showed to be sufficient to reach a convergent result. It uses fully connected layers with a dropout rate of 10\% after each layer to overcome overfitting and to allow for a kind of multiple imputation. In the first hidden layer $l=2$ there are $\phi_2 = 1000$ neurons, as well as in the others except for the output layer. Note that this default choice can be overwritten by the user. 

The layers are applied one after another, while the activations $\mathbf{a}$ of neurons in a first layer determine the activations of the second layer. In a final output layer, the desired imputation of missing values takes place. 
 

Let $\mathbf{X}=\left(\mathbf{x}_{1}, \mathbf{x}_{2}, \ldots, \mathbf{x}_{D}\right)$ be a $n \times D$-dimensional data matrix. $\mathbf{x}_k^{(obs)}$ denotes the observed values of the $k$-th variable, while $\mathbf{x}_k^{(mis)}$ represents the missing values. 
We define two algorithms, while the first one serves as a benchmark when the compositional nature of compositional data is ignored.

\begin{algorithm}[H]
\SetAlgoLined
\KwResult{The imputed data matrix $\mathbf{X}_{(imp)}$ in original scale}
 $\mathbf{d} \leftarrow$ vector of detection limits (one entry for each variable) \;
 $\mathbf{k} \leftarrow$ vector of indices with colums in $\mathbf{X}$ that contains missing values \;
 $\epsilon \leftarrow$ threshold for convergence with sufficient small value, e.g. set to 1 \;
 $m \leftarrow$  number of rounded zeros in $\mathbf{X}$ \;
 $r \leftarrow  0$ index for iterations and $maxiter$ the maximum number of iterations \; 
  Initialization using aKNN (default) or similar methods \;
 \While{$\Delta > \epsilon$ and $r < maxiter$}{
 $r = r + 1$ \;
  $\mathbf{X}^{(imp, old)} \leftarrow$ store previously imputed data matrix\;
  \For{$j$ in $\mathbf{k}$}{
     Fit the deep neural net: $\mathbf{x}_{s} \sim (\mathbf{x}_1,\mathbf{x}_2, ..., \mathbf{x}_D) \setminus \mathbf{x}_s$\ and predict $\mathbf{x}_{s}^{(mis)}$ 
     $\mathbf{X}^{(imp, new)} \leftarrow$ update imputed data matrix with $\mathbf{x}_{s}^{(mis)}$ \;
     Imputed values above $d_s$ are set to $d_s$ \;
    }
    update $\Delta = \frac{1}{m} |\mathbf{X}^{(imp, new)} - \mathbf{X}^{(imp, old)}|$
 }
 \caption{Imputation of rounded zeros with ANNs (without using log-ratios)}
\end{algorithm}

The second algorithm makes use of pivot log-ratio transformations. 
One particular choice of an orthogonal basis using one original composition $(x_{i1}, ..., x_{iD})'$ 
is
\begin{equation}
\label{e3ilr}
z_{ij}=\sqrt{\frac{D-j}{D-j+1}}\,\ln\frac{x_{ij}}{\sqrt[D-j]{\prod_{k=j+1}^D
x_{ik}}} \quad \mbox{ for } \quad j=1,\ldots ,D-1 .
\end{equation}
These coordinates will be referred to 
as \textit{pivot} (log-ratio)
\textit{coordinates} \citep{Filzmoser_2018}.

As mentioned above, the pivot coordinates have the feature that part $\mathbf{x}_{1}$ only
appears in coordinate $\mathbf{z}_{1}$. This is not the case for other parts; $\mathbf{x}_2$, for example,
appears in $\mathbf{z}_1$ and in $\mathbf{z}_2$. Isolating one part into one coordinate is appealing,
since $\mathbf{z}_1$ summarizes now all relative information (log ratios) about $\mathbf{x}_1$.

It is possible
to come back to the original parts (up to a scaling factor) by the inverse pivot (log-ratio) transformation \citep[see also][]{Filzmoser_2018}
\begin{eqnarray}
\label{e3ilrinv} \nonumber
x_{i1}&=&\mathrm{exp}\left(\frac{\sqrt{D-1}}{\sqrt{D}}z_{i1}\right),\\ \nonumber
x_{ij}&=&\mathrm{exp}\left(-\sum_{k=1}^{j-1}\frac{1}{\sqrt{(D-k+1)(D-k)}}z_{ik}+
\frac{\sqrt{D-j}}{\sqrt{D-j+1}}z_{ij}\right), j=2,\ldots ,D-1,  \\ 
x_{iD}&=&\mathrm{exp}\left(-\sum_{k=1}^{D-1}\frac{1}{\sqrt{(D-k+1)(D-k)}}z_{ik}\right),
\end{eqnarray}

The scaling factor is different for each composition in the imputation context. For more details on this factor, we refer to \cite{templ16highdim}.

\begin{algorithm}[H]
\SetAlgoLined
\KwResult{The imputed data matrix $\mathbf{X}_{(imp)}$ }
 Use the same parameters as described in Algorithm 1 \;
 Initialization using aKNN (default) or similar methods \;
 \While{$\Delta > \epsilon$ and $r < maxiter$}{
  \For{$j$ in $\mathbf{k}$}{
  Sort the data matrix so that $\mathbf{x}_j$ is the first column  \;
  $\mathbf{Z}\leftarrow$ apply a pivot log-ratio transformation on the sorted $\mathbf{X}$ \;
  $\mathbf{Z}^{(imp, old)} \leftarrow$ store previously imputed data matrix\;
  $\phi \leftarrow$ represent the detection limit regarding $\mathbf{z}_{1}$ in pivot log-ratio coordinates \;
     Fit the deep neural net: $\mathbf{z}_{1} \sim (\mathbf{z}_2,\mathbf{z}_3, ..., \mathbf{z}_{D-1})$ and predict $\mathbf{z}_{1}^{(mis)}$ 
     $\mathbf{Z}^{(imp, new)} \leftarrow$ update imputed data matrix with $\mathbf{z}_{1}^{(mis)}$ \;
     Imputed values above $\phi$ are set to $\phi$ \;
     Apply the inverse pivot log-ratio transformation \;
     Backsort to the original order of variables \;
     Adjust values so that absolute values are preserved \;
     
    }
    update $\epsilon$
 }
 \caption{Imputation of rounded zeros with ANNs (with log-ratios)}
\end{algorithm}


\section{Numerical results}
\label{numeric}

The numerical results are all based on real data. 
To evaluate the imputation of rounded zeros on real data, detection limits for compositional parts are artificially chosen. This way the imputations can be compared with the real observed values. Note that in principle simulation studies could also be done with artificially generated data. However, it is more realistic to work with real data - where detection limit problems occur in principle - and to set detection limits there. Furthermore, deep learning methods require a lot of computing time (see Figure~\ref{computationtimes}) and therefore it is unrealistic to perform large simulation studies with many replications or varying an extensive grid of parameters for one real data set.

\subsection{Data}
\label{datasets}

\noindent \textit{Kola C-horizon and Moss data:} 
The \emph{Kola Ecogeochemistry Project} was a geochemical survey of the Barents
region whose aim was to reveal the environmental conditions in the European
arctic. Soil samples were taken at different levels and are linked to spatial
coordinates. In this paper, the C-horizon and the Moss data \citep{Rei08} are used. 
The C-horizon data represent soil samples at approximately 1 1/2 meters depth under the surface, the Moss data represent soil samples from the mosses on the surface. In the Moss data, one can see the environmental influence (e.g. pollution by nickel smelters) while in the C-horizon the type of rock has a decisive influence on the soil sample. 
The raw data sets, in which zeros are the result of element concentrations below the detection limit, are included in the R package \texttt{VIM} \citep{templ12VIM, VIMJSS}.
For this study the main elements (Al, Ca, Fe, K, Mg, Mn, Na, P, Si, Ti) are selected for the C-horizon data and for the Moss data we selected the elements Ag, Al, Ca, Cu, Fe, Mg, Na, Ni, Pb, and Si. These elements did not include any values below the detection limit, thus by artificially setting a detection limit, we can compare the true value with the imputed ones.
The number of observations is 606 (C-horizon) and 594 (Moss data).  \\

\noindent \textit{Gemas data:} 
The GEMAS (Geochemical Mapping of Agricultural and Grazing land Soil) project provides geochemical data on agricultural and grazing land soil. It documents the concentration of almost 60 chemical elements at the scale of Europe.

Element concentrations of 2108 samples of agricultural soil (in 0–20 cm depth) and grazing land soil (0–10 cm) are available in package robCompositions \citep{templ11vera, Filzmoser_2018}.  The average sample density was 1 site per 2500 km$^2$.
For our study we used all chemical elements of the data set: Al, Ba, Ca, Cr, Fe, K, Mg, Mn, Na, P, Si, Sr, Ti, V, Y, Zn, Zr.  \\

\noindent \textit{TrondelagO data:}
A regional-scale geochemical survey of O-horizon samples in Nord-Trondelag including the concentrations of nine different plant types. The data set includes 754 observations and we used the chemical concentrations on Ag, Al, As, Ca, Cr, Cu, Fe, Mg, Na, Ni, Pb, and S, that represents those chemical elements that are often used for analysis \citep[see, e.g.,][]{Templ08cluster,Filzmoser_2018}. 

For all data sets, an artificial detection limit was introduced. For each variable in the data set the 0.05 quantile represents the detection limit. This means that for each variable 5\% of the smallest values are set to zero and are subsequently imputed with different methods. 

\subsection{Validation criteria}
\label{val}

The validation criteria defined in the following \citep[see also][]{Martin12, templ16highdim} assume that the complete
data information is available. 
The symbol $^{\ast}$ is used for imputed data or estimators thereof. \\

\noindent \textit{Relative difference between covariance matrices (RDCM):} 
Let $\mathbf{S}=[s_{ij}]$ be the sample covariance matrix of
the original observations in pivot log-ratio coordinates $z_{ij}$ from Equation~(\ref{e3ilr}) (or any other isometric log-ratio coordinates), 
and let $\mathbf{S}^*=[s_{ij}^*]$ denote the sample covariance
matrix computed with the same isometric log-ratio coordinates
where all the rounded zeros have been imputed. A measure of the relative 
difference  between both
covariance matrices \citep{hron10}, based on the
Frobenius matrix norm $\|\cdot\|_F$, 
is
\begin{equation}
\label{diffvariation}
RDCM = \sqrt{\frac{1}{(D-1)^2} \sum\limits_{i=1}^{D-1}
\sum\limits_{j=1}^{D-1}
\left(s_{ij} - s_{ij}^*\right)^2}=
\frac{1}{D-1}\frac{\|\mathbf{S}-\mathbf{S}^*\|_F}{\|\mathbf{S}\|_F} \quad .
\end{equation}

\noindent \textit{Compositional error deviation (CED):} 
Let $n_M$ denotes the number of samples $\mathbf{x}_k$ containing at least one rounded zero, 
$M$ the index set referring to such samples, and 
$\mathbf{x}^*_k$ the corresponding observation with imputed values. 
Then the criterion

\begin{equation}
\label{CoErDe}
\frac{\frac{1}{n_M}\sum\limits_{k \in M}d_a(\mathbf{x}_k,\mathbf{x}^*_k)}{   \max\limits_{ \{ \mathbf{x}_i,\mathbf{x}_j \in \mathbf{X} \} } \{ d_a(\mathbf{x}_i,\mathbf{x}_j) \}}
\end{equation}
measures the differences by using Aitchison distances (defined below). $d_a$ stands for the
Aitchison distance, which is defined for two compositions $\mathbf{x}$ and
$\mathbf{\tilde{x}}$ as
$d_A(\mathbf{x},\mathbf{\tilde{x}})=\left[\frac{1}{D}\sum_{i=1}^{D-1}\sum_{j=i+1}^D
\left(\log\frac{x_i}{x_j}-\log\frac{\tilde{x}_i}{\tilde{x}_j}\right)^2\right]^{1/2} $
\citep{aitchison00}. This criterion represents a generalization of the measure defined in \cite{hron10} and it was used in \cite{Martin12} and \cite{templ16highdim}. Note that the denominator is 
the maximum distance in the original data set. \\

\noindent \textit{Curious Imputations:}
The number of imputations below 0 and above the detection limit represents the imputation of rounded zeros that are not valid. 

\subsection{Results}
\label{subresults}

\begin{figure}
\begin{center}
\includegraphics[width = 0.95\textwidth]{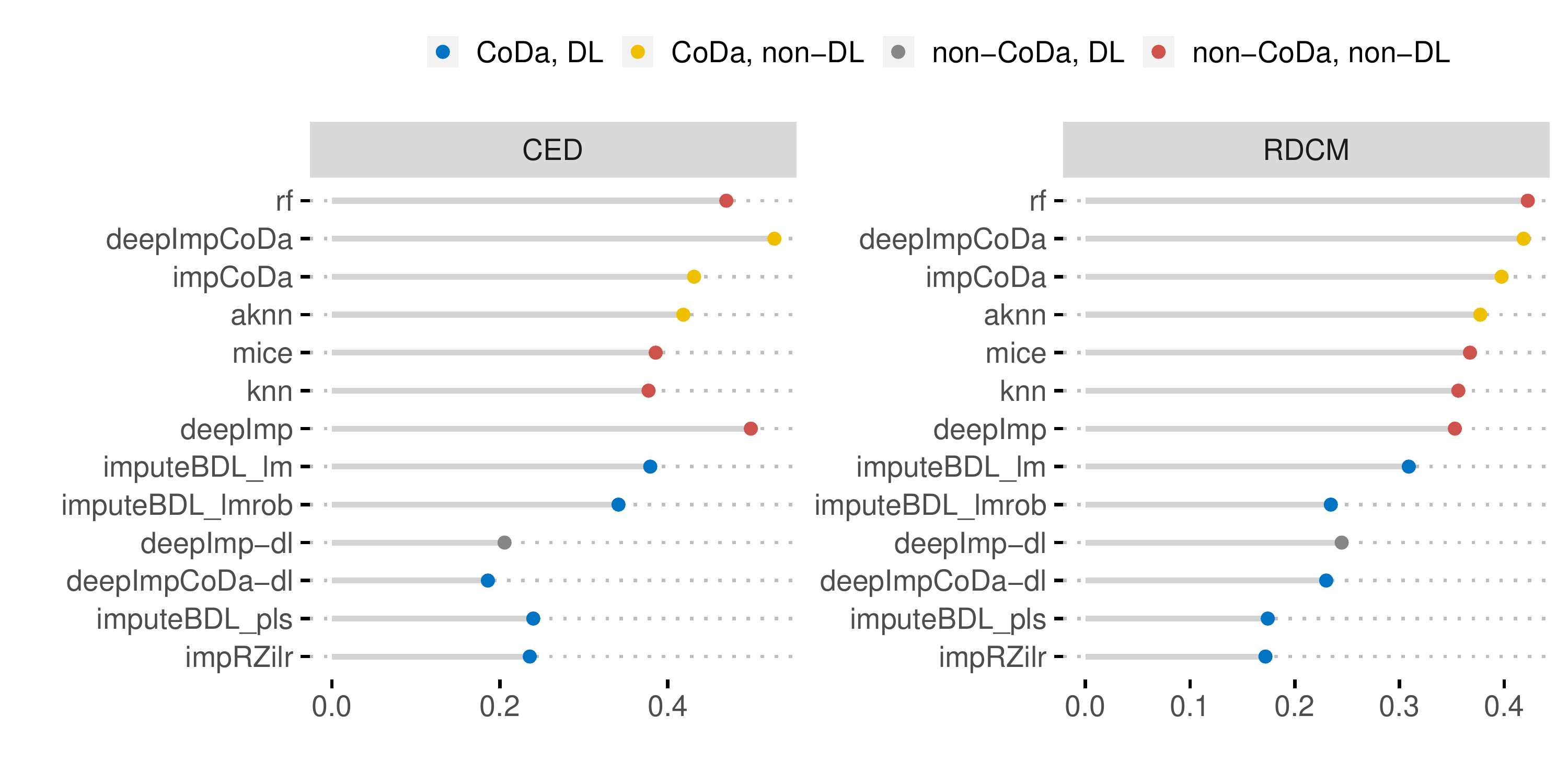}
\caption{\label{comp1}Comparison of methods for the C-horizon data. Methods that take the compositional nature of the data and detection limits into account are shown in blue. The legend expresses the nature of the method. For example, \textit{CoDa, DL} means the use of a compositional method that is designed to deal with detection limits. See Table 1 for details.}
\end{center}
\end{figure}

Looking at the result for the imputation of the C-horizon data in Figure~\ref{comp1}, it can be seen that, except for \texttt{deepImp-dl}, the best-judged methods are those which take the compositional nature of the data and the special problems for rounded zeros into account. 
The very best methods for the imputation of rounded zeros for the C-horizon data are \texttt{impRZilr}, \texttt{imputeBDL\_pls} and \texttt{deepImpCoDa-dl}. Since \texttt{impRZilr} and \texttt{imputeBDL\_pls} differ algorithmically only slightly and both methods represent an EM-algorithm with partial least squares regression, it is not surprising that almost identical results are obtained. Slightly better than \texttt{deepImp-dl}
is \texttt{deepImpCoDa-dl}. This shows that a pivot log-ratio transformation does bring a slight improvement even when using ANNs. The two methods are also best for the evaluation criterion CED , which expresses normalized Aitchison distances between observed and imputed values. Comparatively bad are methods that represent pure imputation methods without taking the problem of rounded zeros into account. The benchmarks \texttt{deepImpCoDa} and \texttt{deepImp} are comparably worse than their rounded zeros counterparts. 
Surprisingly, \texttt{knn} scores slightly better than \texttt{aknn}.

\begin{figure}
\begin{center}
\includegraphics[width = 0.95\textwidth]{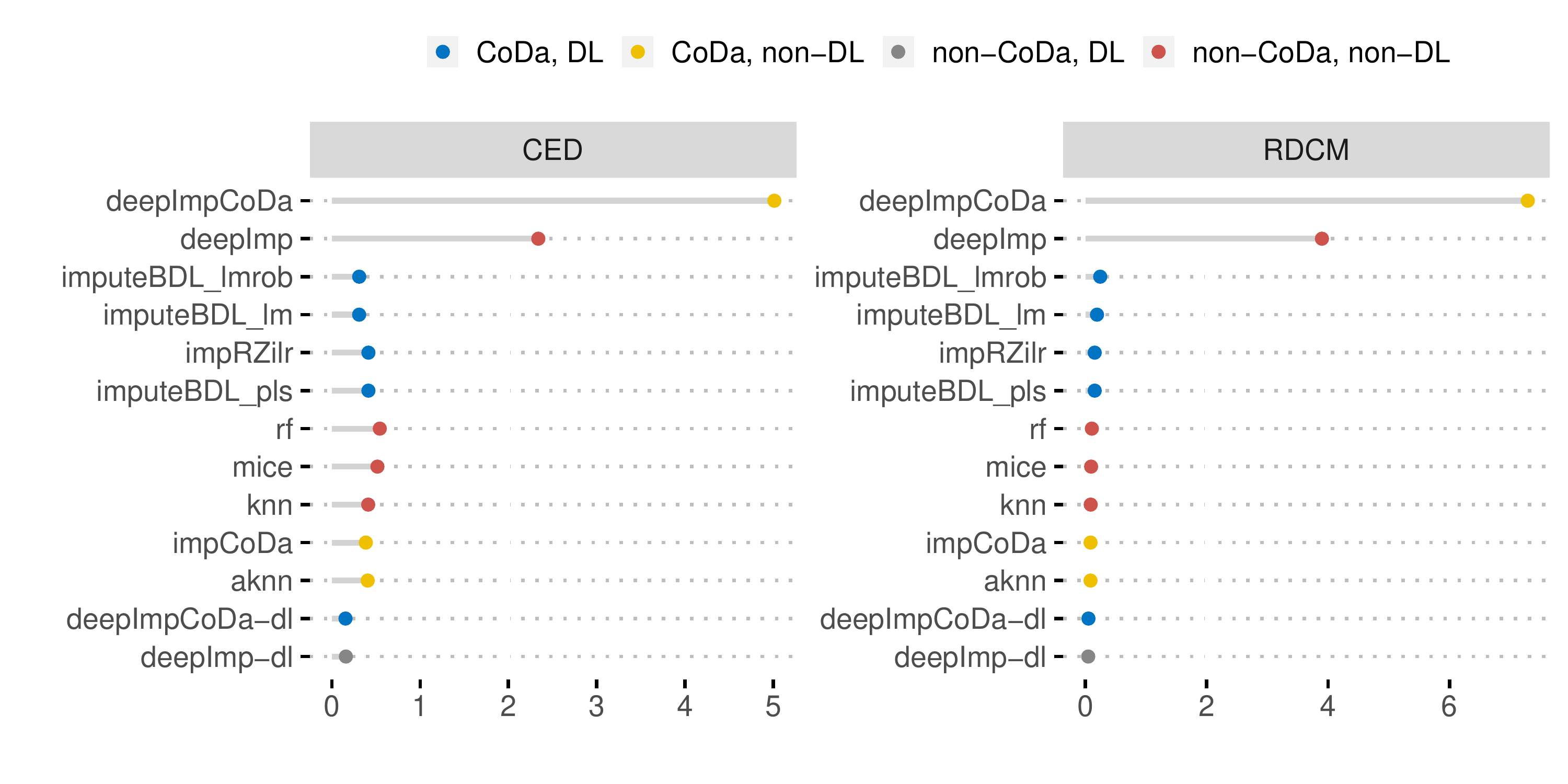}
\caption{\label{comp2}Comparison of methods for the Moss data. Methods that take the compositional nature of the data and detection limits into account are shown in blue.}
\end{center}
\end{figure}

The results of the Moss data are shown in Figure~\ref{comp2}. 
The two new methods, \texttt{deepImpCoDa-dl} (ced: 0.1539, RDCM: 0.0516) and \texttt{deepImp-dl} (ced: 0.1584, RDCM: 0.0479) are by far the best, followed by \texttt{aknn}. (ced: 0.4053, RDCM: 0.0856). So there is no difference whether algorithm 1 or algorithm 2 is used with respect to ANNs, i.e. the deep neural network learns the restrictions of the simplex indirectly and also maps those possibly non-linear relationships which may become linear or better represented by a log-ratio transformation. 
The other methods for rounded zeros also perform relatively poorly, for example \texttt{imputeBDL\_pls} (ced: 0.4136, RDCM: 0.1545).
The benchmark methods \texttt{deepImp} and \texttt{deepImpCoDa} work very poorly, neither of which addresses the problem of detection limits. 

A very similar picture can be seen in the results for the TrondelagO data in Figure~\ref{comp3}. Again, the two algorithms 1 (Benchmark: \texttt{deepImp-dl}, CED: 0.2608, RDCM: 0.1645) and algorithm 2 (CoDa-Adaption: \texttt{deepImpCoDa-dl}, CED: 0.2539, RDCM: 0.1642) performed best by far. Again, a representation in pivot log-ratio coordinates is not important and the deep neural network learns the relationships even in simplex. Compared to other methods, the rounded zeros methods also perform relatively well in the CED criterion, for example \texttt{imputeBDL\_pls} (ced: 0.6508, RDCM: 0.2915) is better than for example mice (CED: 0.9552, RDCM: 0.2708). The two benchmarks \texttt{deepImp} and \texttt{deepImpCoDa} are comparatively poor. The reason is that a few imputed values were imputed very poorly.  

\begin{figure}
\begin{center}
\includegraphics[width = 0.95\textwidth]{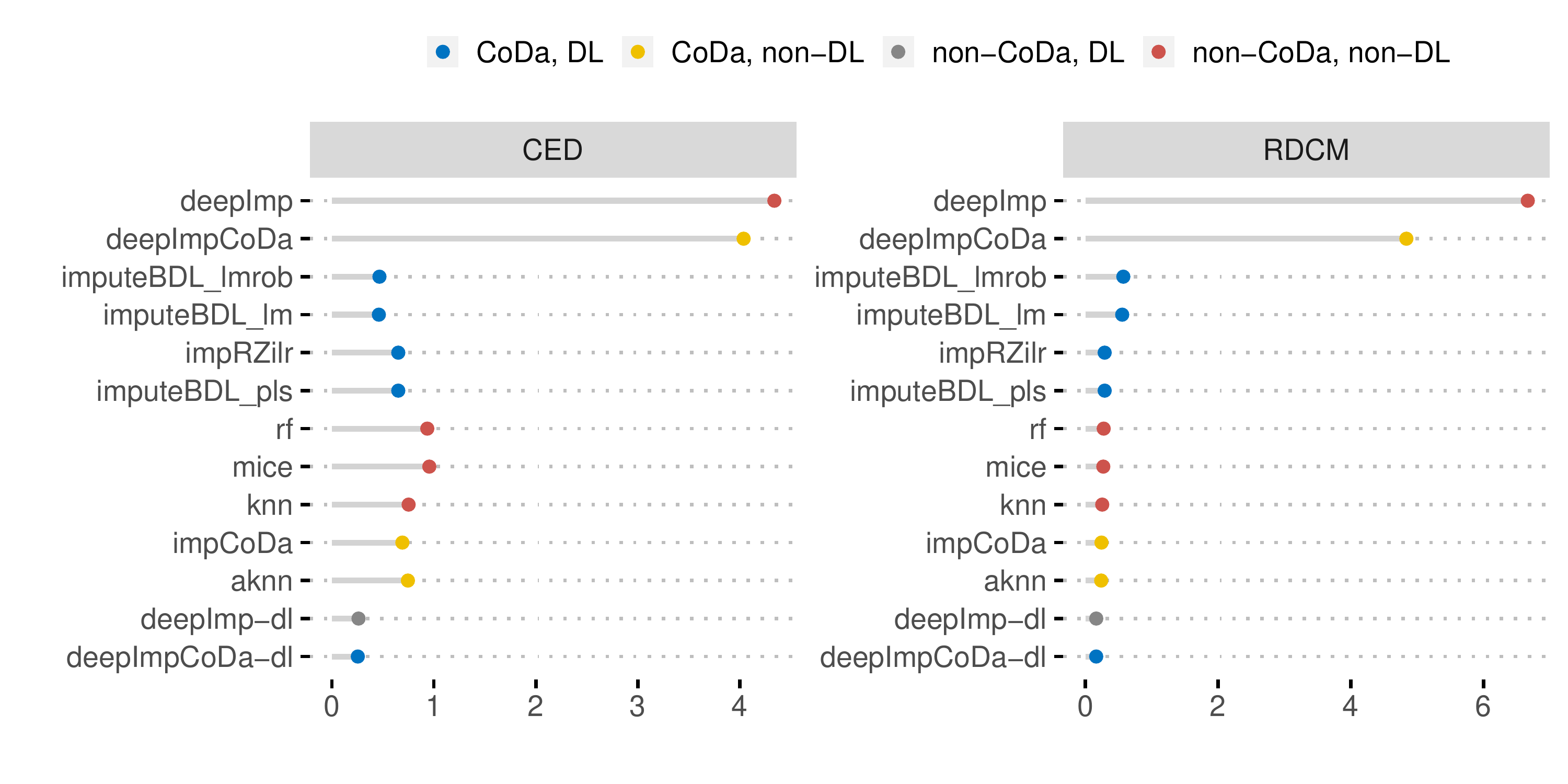}
\caption{\label{comp3}Comparison of methods for the TrondelagO data. Methods that take the compositional nature of the data and detection limits into account are shown in blue.}
\end{center}
\end{figure}

By far the best rounded zero imputation method for the Gemas data is Algorithm 2 (\texttt{deepImpCoDa-dl}, CED: 0.1721, RDCM: 0.0971), followed by Benchmark Algorithm 1 (\texttt{deepImp-dl}, CED: 0.2538, RDCM: 0.1195). 
All other methods are comparatively poor, for example \texttt{imputeBDL\_pls} (CED: 0.5201, RDCM: 0.5172). It is interesting that for the ANNs this time the pivot log-ratio transformations contribute to the improvement of the results. This means in this case that the ANNs did not learn the compositional structure of the data, but an isometric log-ratio transformation is needed to improve the results (compare \texttt{deepImpCoDa-dl} with \texttt{deepImp-dl} and \texttt{deepImpCoDa} with \texttt{deepImp}). 

\begin{figure}
\begin{center}
\includegraphics[width = 0.95\textwidth]{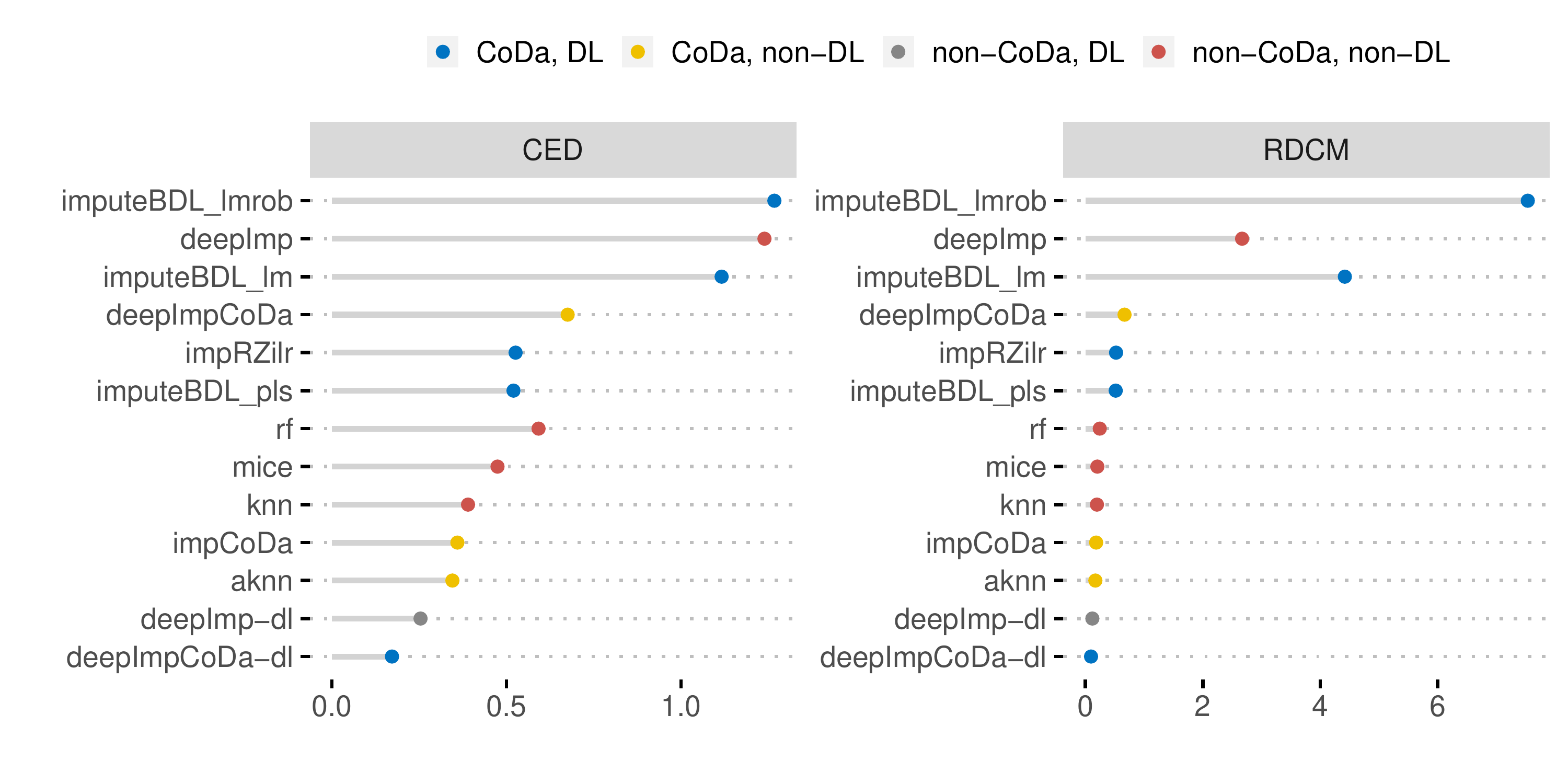}
\caption{\label{comp3}Comparison of methods for the Gemas data. Methods that take the compositional nature of the data and detection limits into account are shown in blue.}
\end{center}
\end{figure}

\begin{figure}
\begin{center}
\includegraphics[width = 0.95\textwidth]{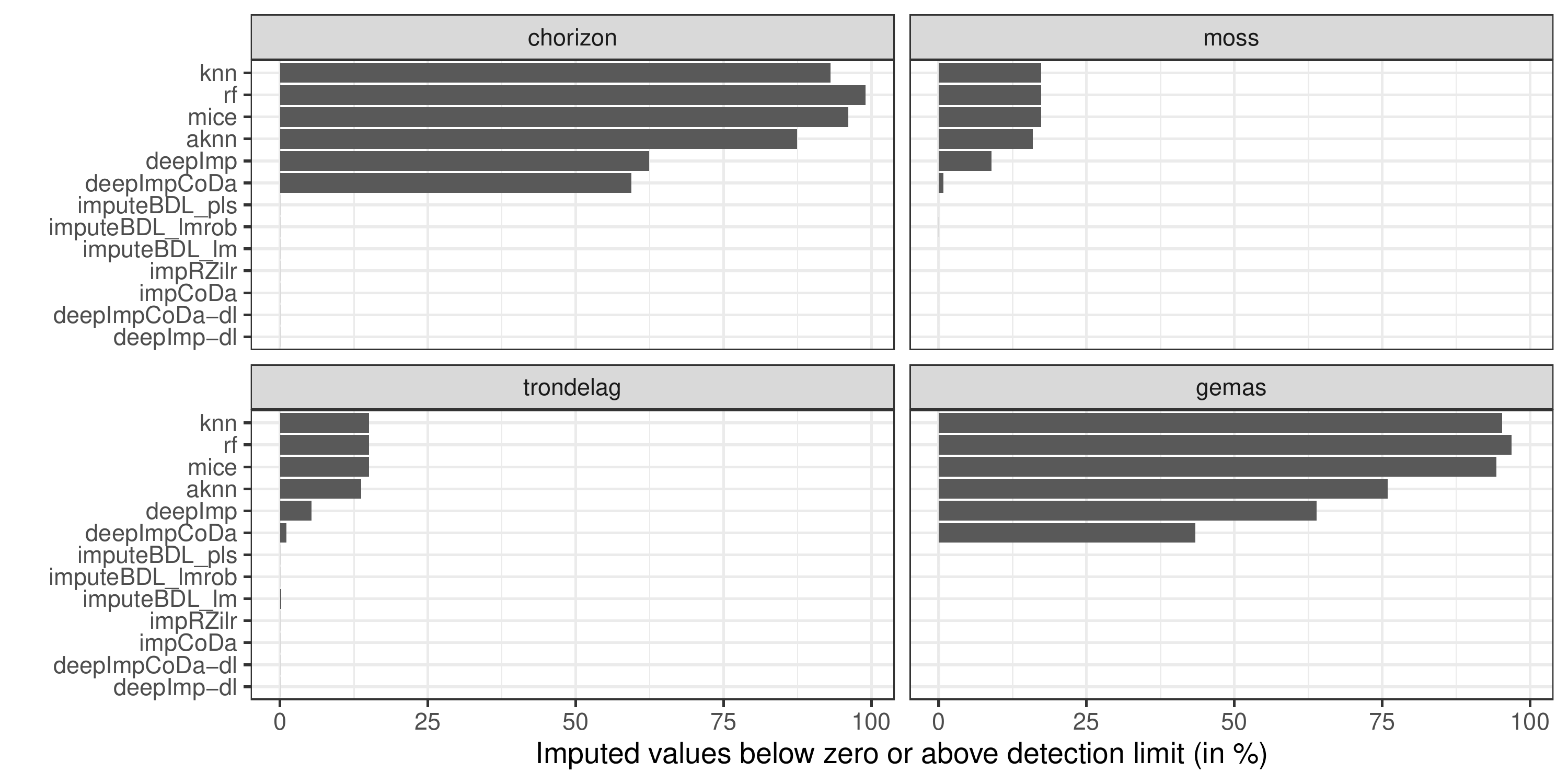}
\caption{\label{outsiders}Percent of imputed values above the detection limit or below 0.}
\end{center}
\end{figure}

In Figure~\ref{outsiders} one can see which imputation methods  do not take detection limits into account and very often they impute values outside the interval (0, detection limit). While this is not such dramatic with the Moss and TrondelagO data, it is evident with the C-horizon and Gemas data. It is interesting to note that \texttt{deepImpCoD} imputes comparatively fewer values outside the intervals than \texttt{deepImp}, since no values below 0 can be imputed  with this method. The use of methods that do not take detection limits into account is generally critical, since many values outside the definition range, censored with 0 and detection limit, are imputed. 
Note that values outside the interval (0, DL] can be shrinked to the limits. This means that imputed values above the detection limit are winsorized to the detection limit, and values less than or equal to zero are set to zero + constant, where typically the constant is often chosen very small. The latter implies that outliers may be generated by a log-ratio transformation.

%
%
%
%
%

\subsection{Computation time}

\begin{figure}
\begin{center}
\includegraphics[width = 0.95\textwidth]{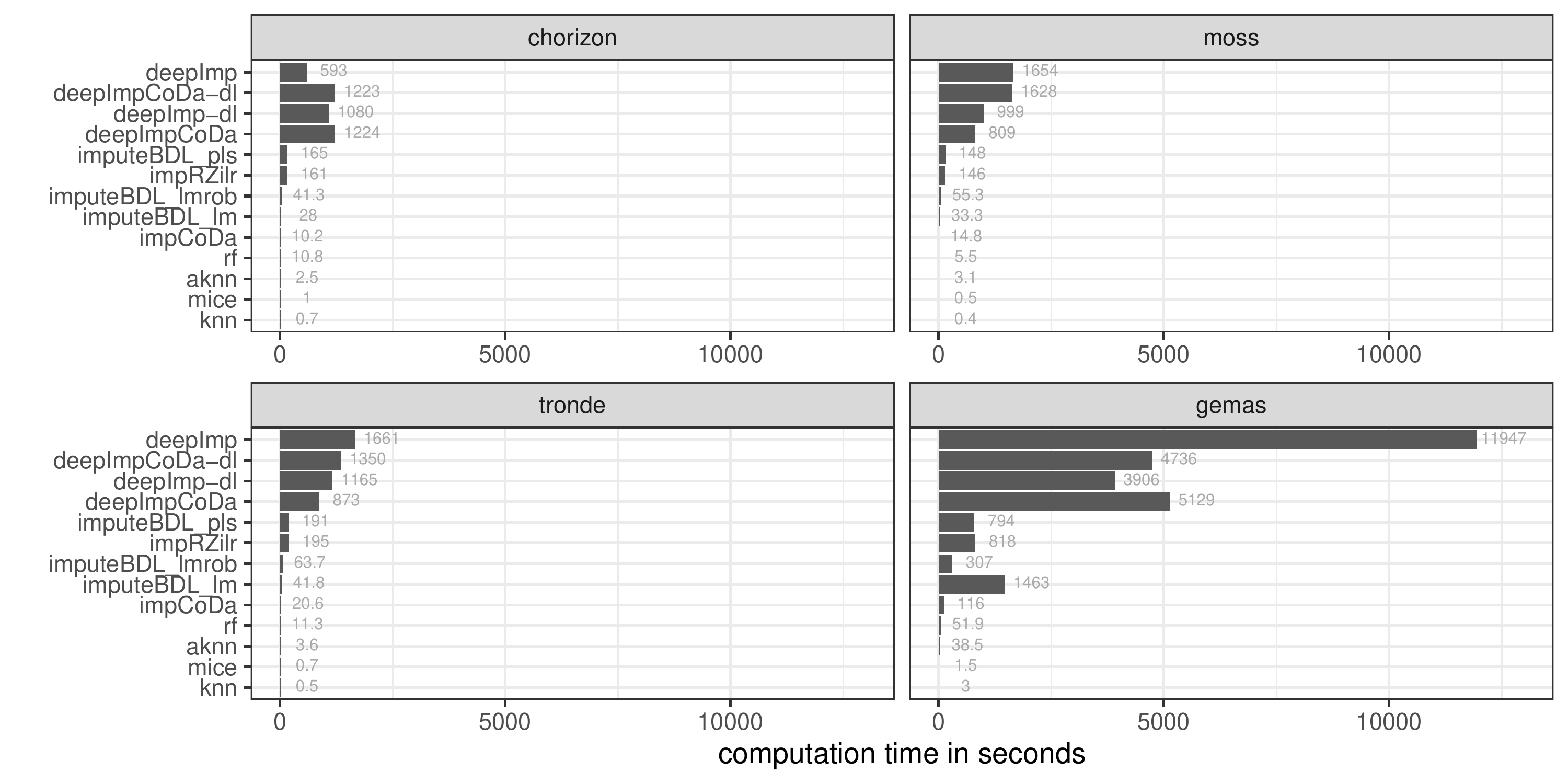}
\caption{\label{computationtimes}Computation times on all data sets.}
\end{center}
\end{figure}

The computational cost of our ANN imputation framework is compared with the runtimes of other imputation procedures using a 2.6 GHz Intel Core i7 single cpu in Table~\ref{computationtimes}. The kNN method implemented in package VIM \citep{templ12VIM, VIMJSS} is the fastest in general, however, this method does not consider the rounded zero problem adequately. The runtimes of the EM-based methods using pivot coordinates in \cite{Martin12, templ16highdim} are computationally faster than the artificial neural network methods by approximately a factor of 10. Note that \texttt{imputeBDLs\_rob} should be generally faster than \texttt{imputeBDLs\_lm}, but in the EM-based approach, the linear model approach takes longer to converge as with the robust method, i.e. the imputations need longer to stabilize and more iterations are necessary. This is also the case with \texttt{impNNet}.

The increase of the quality of the imputation using our ANN framework for the imputation of rounded zeros has a price regarding the computation time.

However, if needed, there are several ways to speed up the computation of our ANN framework, namely to use fewer layers, fewer neurons per layer, fewer epochs, and/or fewer iterations.

\section{Conclusion}
\label{conclusions}

Up to our knowledge and literature research on the date of the submission of this paper, this is the first peer-reviewed contribution in compositional data analysis with ANN's and, more general, using deep learning approaches. 
The proposed new methods for the imputation of rounded zeros were compared not only with other methods for the imputation of rounded zeros but also with popular conventional imputation methods. As already mentioned, this sounds a bit strange at first glance, since rounded zeros are exclusively censored data. The motivation for this comes from practical and subjective experience as package maintainer of imputation packages and author of many imputation functions. This subjective experience shows that many users and researchers haven't understood that the problem of rounded zeros requires specially designed methods and that users and researchers mostly make the wrong choice of imputation methods, although the articles and function descriptions are clear on the choice of methods. 
Also, with such a comparison we highlight the compositional data problem to research areas where compositional data are not well known. 
For this reason, we will compare exemplary imputation methods with general imputation methods to make this more obvious and evident. 
Note that a practitioner who neglect compositional data analysis methods could criticize that the compositional data methods are probabably only better because of the suggested compositinal evaluation measures. However, compositional data also require compositional validation critera. 
Furthermore, the results shown in Figure~\ref{outsiders} are a strong argument against the use of non-compositional ``classical'' imputation methods because these methods are often used to impute above the detection limit. 

Also, benchmark methods were determined. For example, Algorithm 1 is not adequate from a theoretical point of view, but it is interesting to compare it as a benchmark for Algorithm 2 to answer the question of how far results improve on real data when using log-ratio transformations. 

The newly introduced method using ANNs (\texttt{deepImpCoDa-dl}) is not only competitive but consistently gave the best results for all three data sets. The Moss data has better correlations between the variables than the C-horizon data - this gives a boost to the ANNs. The TrondelagO data have a good group structure concerning different plant species, which the ANNs learn better than other methods. The Gemas data are significantly larger than the other data sets, which has an impact on the quality of the ANNs, which generally improve as more training, testing, and validation data are available. The presented imputation methodology with ANNs beats all other methods by far. Note that for other data sets it is likely to reveal different measures of performance caused by distinct multivariate correlation and dependency structures. 

It is interesting to see that a representation in (pivot) log-ratio coordinates is not necessarily needed anymore, but the results hardly deteriorate if the data are not represented in log-ratio coordinates, because the limitations of the simplex and the coherences in the data are well learned in the deep neural network. Theoretically, however, the correct representation of the data also plays a role in deep neural networks, since the distances between observed values should be considered in a compositional way. We also see differences in the C-horizon results and especially in the results for the Gemas data set, where Algorithm 2 (with log-ratio transformations) performs better than the benchmark Algorithm 1 (without log-ratio transformations).  

A drop of bitterness is the long computing times because an ANN with enough layers and neurons takes a long time to train. For typical data sets in geochemistry, for example, this is not a problem as long as no parameter optimization on the number of layers, neurons, loss-function, evaluation-function, activation function, epochs, etc. is made. 
In practice, the imputation is often performed exactly once, and therefore it doesn't matter if an algorithm is ready in a second or if it takes 4 hours. In any case, we provide sensible defaults and stop whenever the evaluation criteria will not improve after a given number of epochs. Multiple imputation is possible in principle and can be controlled by the parameter \texttt{dropout}, which also can avoid overfitting. 

The presented algorithms using ANNs are part of a larger R package for deep learning in imputation and compositional data analysis that is work in progress, and it will be published independently of this contribution.

We propose three topics for future research. 
Firstly, it would be interesting to compare the methods from the R package \textit{zCompositions}. Although these methods are mostly of univariate nature or include a multiple EM-based method based on the alr transformation, in many cases rounded zeros are very well imputed. Secondly, large simulation studies may, under certain circumstances, lead to further knowledge about the quality of the methods.  Thirdly, in this contribution we only looked at some well-known evaluation measures such as the CED and RDCM but further analysis is possible. For example the evaluation of classifiers on the imputed data can show which methods are more advantageous and lead to fewer missclassifications, and log-ratio biplots or comparison of results from a regression model can show what the effects of imputing rounded zeros are. 

\begin{acknowledgement}
I would like to thank Peter Filzmoser and Karel Hron for the many collaborations and the fruitful discussion on the topic of compositional data analysis, imputation, and rounded zeros. Furthermore, my thanks go to Eric Grunsky and Peter Filzmoser as well as to one unknown reviewer for their constructive and helpful comments on the initial submission. 

\end{acknowledgement}

\bibliographystyle{plainnat}
\bibliography{annbib}

\end{document}